%% file: main.tex
\documentclass[letterpaper, 10 pt, conference]{ieeeconf}  %

\IEEEoverridecommandlockouts                              %

\usepackage{amsmath} %
\usepackage{amssymb}  %
\usepackage{graphicx}
\usepackage{csquotes}
\usepackage[english]{babel}
\usepackage[hidelinks]{hyperref}
\usepackage{balance} %
\usepackage{cite}
\usepackage{soul,color}
\usepackage{float}
\usepackage[dvipsnames]{xcolor}
\usepackage{siunitx}

\usepackage{booktabs}
\usepackage{multirow}

\usepackage{booktabs} %
\usepackage{tabularx}
\usepackage{multirow}

\usepackage{subfig}
\let\labelindent\relax
\usepackage{enumitem} %

\usepackage[final]{changes}
\setdeletedmarkup{\textcolor{red}{\sout{#1}}}

\usepackage{todonotes}
\newcommand{\virgolette}[1]{``#1''}

\newcommand{\fref}[1]{Fig.~\ref{#1}}
\newcommand{\sref}[1]{Sec.~\ref{#1}}

\newcommand{\tref}[1]{Table~\ref{#1}}

\addto\captionsenglish{} %

\title{\LARGE \bf
Wearable Haptics for a Marionette-inspired Teleoperation \\ of Highly Redundant Robotic Systems
}

\author{Davide Torielli$^{1,3}$, Leonardo Franco$^{2}$, Maria Pozzi$^{1,2}$, Luca Muratore$^{1}$, \\ Monica Malvezzi$^{2}$, Nikos Tsagarakis$^{1}$, and Domenico Prattichizzo$^{1,2}$%
\thanks{*We gratefully acknowledge the funding provided by the EU Horizon Europe projects HARIA (GA No.101070292) and SOPHIA (GA No.871237)}
\thanks{$^{1}$Humanoids \& Human Centered Mechatronics Research Line, Istituto Italiano di Tecnologia, Genova, Italy (\{name.surname\}@iit.it)}%
\thanks{$^{2}$Department of Information Engineering and Mathematics, University of Siena, Siena, Italy 
(\{name.surname\}@unisi.it).}
\thanks{$^{3}$Department of Informatics, Bioengineering, Robotics, and Systems Engineering (DIBRIS), University of Genova, Genova, Italy}%
}

\begin{document}
	
    \maketitle
    \thispagestyle{empty}
    \pagestyle{empty}

\input{sez00_abstract}

\section{Introduction}

\input{sez01_intro_MP}

\section{Related works}\label{sec:relwork}
\input{sez02_soa}

\section{Methodology}\label{sec:implementation}
\input{sez03_implementation}

\section{Experiments}\label{sec:ris}

\input{sez04_experiments}

\section{Results and Discussion}\label{sec:disc}

\input{sez05_discussion}

\section{Conclusions}\label{sec:conclusion}

\input{sez06_conclusion}

\addtolength{\textheight}{0cm}   %
\balance

\bibliographystyle{IEEEtranBST/IEEEtran}
\bibliography{IEEEtranBST/IEEEabrv,bib.bib}

\end{document}

%% file: sez00_abstract.tex
\begin{abstract}

The teleoperation of complex, kinematically redundant robots with loco-manipulation capabilities represents a challenge for human operators, who have to learn how to operate the many degrees of freedom of the robot to accomplish a desired task. In this context, developing an easy-to-learn and easy-to-use human-robot interface is paramount. Recent works introduced a novel teleoperation concept, which relies on a virtual physical interaction interface between the human operator and the remote robot equivalent to a \virgolette{Marionette} control, but whose feedback was limited to only visual feedback on the human side. 
In this paper, we propose extending the \virgolette{Marionette} interface by adding a wearable haptic interface to cope with the limitations given by the previous work\added{s}. Leveraging the additional haptic feedback modality, the human operator gains full sensorimotor control over the robot, and the awareness about the robot's response and interactions with the environment is greatly improved.
We evaluated the proposed interface\deleted{,} and the related teleoperation framework\deleted{,} with naive users, assessing the teleoperation performance and the user experience with and without haptic feedback. The conducted experiments consisted in a loco-manipulation mission with the CENTAURO robot, a hybrid leg-wheel quadruped with a humanoid dual-arm upper body.
	
\end{abstract}

%% file: sez01_intro_MP.tex
Highly redundant robots with complex kinematic structures (e.g., humanoids, leg-wheel platforms, multi-arm systems) are difficult to teleoperate. Many approaches have been proposed to solve this challenge, including strategies adopting shared control techniques~\cite{Selvaggio2021, TPO2}, or designing human-robot interfaces that go beyond classical joysticks, exploiting novel Body-Machine Interfaces (BoMI)~\cite{Casadio2012,Gasper2021,TPO}, and establishing a multi-modal feedback connection between the two agents~\cite{pacchierotti2015cutaneous,dafarra2022}. 
However, a teleoperation interface capable of accounting for multiple inputs from the user and feeding back different kinds of stimuli might easily compromise ergonomics and become impractical.

\begin{figure}
	\centering
	\includegraphics[width=0.95\linewidth]{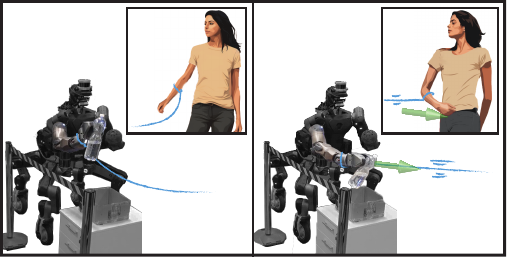}
	\caption{Concept: On the left, the blue rope, representing the TelePhysicalOperation connection, is not under tension. The robot stands still when the operator relaxes her arm. On the right, as soon as the user draws her arm closer to her body, the blue rope increases its tension, pulling the robot arm that follows her movement, while she can perceive the force feedback of pulling through our novel framework.
      \vspace{-7px}
    }
	\label{fig:intro}
\end{figure}

This paper addresses the trade-off between interface complexity and usability by building a new wearable sensorimotor interface that adds the haptic feedback channel and new control inputs to the system proposed in a previous work by the authors~\cite{TPO}. 
The adopted technology is lightweight and unobtrusive, completing the TelePhysicalOperation (TPO) framework, in which the robot is commanded through virtual ropes attached to specific robot body parts that can be pushed and pulled by the operator's arm motion (\fref{fig:intro}). 
While in the initial implementation of the system the user had only limited visual feedback of the robot and its workspace, in this work, we exploit cutaneous haptic cues to provide the user with further information. In addition, we integrate push buttons to allow the user to send additional control inputs through the TPO framework. 
The resulting novel haptics-enabled TPO human-robot interface allows the full sensorimotor interaction~\cite{prattichizzo2021human} between the human and the robot, providing also flexibility to be adapted to different teleoperation scenarios.

The haptic interface consists of two rings and two armbands, which can deliver vibration and skin indentation stimuli. The rings are also endowed with two small push buttons. The user wears a ring and an armband on each upper limb. 
In the proposed implementation, skin indentation stimuli are sent to the user to increase his/her awareness of the control command delivered to the robot and of the robot-environment interaction, whereas vibrations are used as acknowledgments about the execution of the command associated with a button. While the benefits of vibratory acknowledgments and force feedback in teleoperation have been shown in previous works~\cite{pacchierotti2015cutaneous, Franco2019}, here the haptic cues are specifically tailored to the TPO system. In particular, a new skin indentation feedback strategy that conveys the sensation of pulling the virtual ropes has been designed. Whenever a robot part is being controlled by the operator\added{'s arm} and is actually moving, the corresponding operator's forearm is squeezed proportionally to the entity of the control action.

To summarize, the main contributions of this work are:
\textit{(i)} The development of a haptic feedback strategy designed for the \enquote{Marionette}-inspired TPO framework~\cite{TPO};
\textit{(ii)} The hardware development of an advanced sensorimotor interface combining input (tracking cameras, buttons) and feedback (haptic rings and armbands) devices and its integration in the TPO system;
\textit{(iii)} The evaluation of the overall interface with naive operators who teleoperated the CENTAURO robot~\cite{Centauro2}, a complex leg-wheel platform with a humanoid dual-arm upper body, in a loco-manipulation mission.

%% file: sez02_soa.tex
The growing complexity of robots and the effective exploitation of their enhanced capabilities can be tackled by the development of novel intuitive human-robot interfaces. These interfaces must minimize the operators' learning curve and augment their effectiveness during the task execution~\cite{Gaofeng2023}. In pursuit of this goal, researchers have focused on surpassing the conventional interaction means, such as joysticks~\cite{VILLANI2018}, by developing innovative BoMI, to let the operator control the robot with his/her own body~\cite{Casadio2012}. Body gestures have been tracked and mapped to specific robot commands with a range of technologies, including Inertial Measurement Units~\cite{Gasper2021}, motion capture~\cite{nunez2018}, and Electromyography~\cite{Vogel2011}. %

Together with the implementation of more intuitive control interfaces, it is crucial to provide operators with a deeper understanding of the events unfolding at the remote robot location. Substantial efforts have been dedicated to developing devices that can transmit a heterogeneous range of sensory information to the user, which is often essential to work with high-Degrees Of Freedom (DOF) robots~\cite{Darvish23}. Indeed, the direct observation of the robot or the acquisition of information through a monitor does not always guarantee sufficient insights about the tasks or the robot state, neither it provides an immersive experience to the operator~\cite{MONIRUZZAMAN2022}.
Consequently, several works have coupled the remote robot control with a sensory feedback channel that supplements the visual domain, encompassing an array of stimuli like haptic cues to leverage the human sense of touch~\cite{Dargahi2004, Pacchierotti2015}. 
Various types of tactile devices to control diverse robotic platforms have been proposed, spanning from wheeled robots~\cite{Zhao2023}, to underwater vehicles~\cite{XIA2023}, drones~\cite{Macchini2020}, and humanoids~\cite{Schwarz2021,dafarra2022,baek2022}. %
Wearable interfaces, in particular, allow \replaced{delivering}{to deliver} a variety of cutaneous cues (e.g., skin stretch, vibration, temperature) to different body parts, resulting in unobtrusive, flexible devices capable of transmitting different kinds of information to the user~\cite{pacchierotti2017wearable}. 
For example, in~\cite{Brygo2014}, a vibrotactile belt alerts the operator about the loss of balance of a bipedal robot, whereas in~\cite{Bimbo2017,BAI2019}, vibrotactile patterns\added{,} delivered to the user hand\added{,} aid the operator in guiding a robot avoiding obstacles.
Nevertheless, it is still a challenge to develop a wearable haptic interface that allows the control of complex robots while maintaining a simple BoMI. Indeed, some solutions proposed in the literature offer extensive inputs and feedback possibilities, but are cumbersome to wear~\cite{Brygo2014,Schwarz2021,dafarra2022,baek2022}; other approaches limit the flexibility with grounded haptic devices~\cite{Pacchierotti2015, pacchierotti2015cutaneous, BAI2019, Correa2022, Cheng2022}, or with an external tracking system~\cite{nunez2018}; while unobtrusive wearable interfaces are usually adopted to control robots with a lower number of DOFs~\cite{Vogel2011,Bimbo2017,Gasper2021}.

%% file: sez03_implementation.tex
\begin{figure*}
	\centering
	\includegraphics[width=0.9\linewidth]{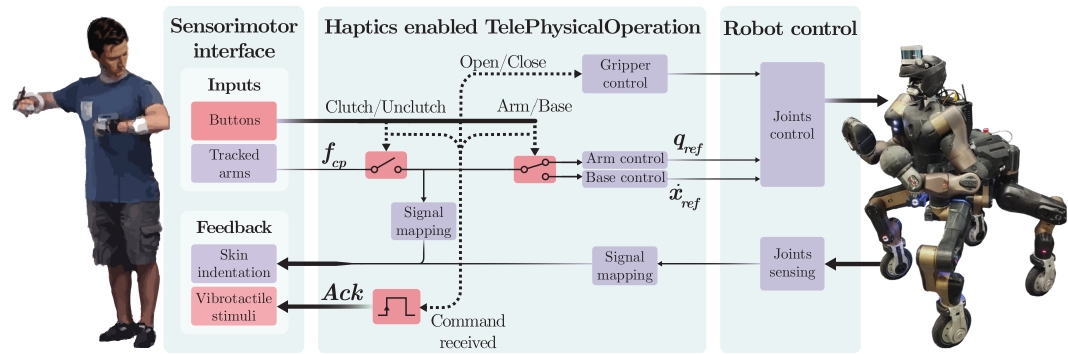}
    \caption{Scheme of the proposed haptics-enabled TelePhysicalOperation framework.}
	\label{fig:architecture}
\end{figure*}

The proposed teleoperation framework (\fref{fig:architecture}) is based on the TelePhysicalOperation (TPO) paradigm enhanced with the newly presented haptic interface. 

\subsection{TelePhysicalOperation Control}\label{sec:TPO}
In our previous work, we presented the TPO concept~\cite{TPO}, a novel teleoperation interface inspired by the \enquote{Marionette} motion control principle. With this interface, the operator pulls and pushes virtual ropes applying virtual forces on selected robot body segments from a remote location. These virtual forces are generated by the movements of the user's arms, tracked by tracking cameras worn on the \deleted{person's }wrists. Based on the direction and magnitude of the virtual forces, the robot \deleted{will }generate\added{s} corresponding motions, as if they were real forces applied during a physical human-robot interaction. In this way, it is possible to control the robot posture, maintaining the intuitiveness of physical human-robot interaction, but also remotely controlling the robot, avoiding potential safety issues\added{ introduced by the physical interaction between the user and the robot}. The virtual ropes of the \enquote{Marionette} interface can be attached to different robot body parts, such as the arm(s) and the mobile platform body, allowing to control the manipulation and mobility of a mobile manipulation robot.

Specifically, each user's arm displacement generates a virtual force $\boldsymbol{f}_{cp} \in \mathbb{R}^{3\times1}$, imposed on a selected robot control point (\textit{cp}). Given an N-joint manipulator, the joint position reference vector $\boldsymbol{q}_{ref}(t) \in \mathbb{R}^{N\times 1}$ is computed by a joint-level admittance control law as follows:
\begin{equation*} 
    \boldsymbol{\ddot{q}}_{ref}(t) = \boldsymbol{M}^{-1} \big ( \boldsymbol{K} (\boldsymbol{q}_{eq} - \boldsymbol{q}(t)) - \boldsymbol{D} \boldsymbol{\dot{q}}_{ref}(t-1)+\boldsymbol{J}^T \boldsymbol{f}_{cp} \big),
\end{equation*}
where $\boldsymbol{M}, \boldsymbol{K}, \boldsymbol{D} \in \mathbb{R}^{N\times N}$ are the diagonal matrices of the mass, stiffness, and damping parameters of the joint mass-spring-damper model;  $\boldsymbol{q} , \boldsymbol{q}_{eq} \in \mathbb{R}^{N\times 1}$ are the current position of the joints and the equilibrium set point where a stiffness greater than zero will drag the joints; $\boldsymbol{\ddot{q}}_{ref}$ is integrated twice to obtain the joint position reference $\boldsymbol{q}_{ref}$. 

Given a mobile base robot, the mobility ability is controlled by generating a Cartesian velocity reference $\boldsymbol{\dot{x}}_{ref} \in \mathbb{R}^{3\times1}$ from the virtual force:
	$\boldsymbol{\dot{x}}_{ref} = \boldsymbol{K}_{cart,cp} ~\boldsymbol{f}_{cp},$ 
where $\boldsymbol{K}_{cart,cp} \in \mathbb{R}^{3\times 3}$ is a diagonal matrix of gains. Further details about the TPO way of control can be found in~\cite{TPO}.

On the right side of \fref{fig:architecture}, from $\boldsymbol{\dot{x}}_{ref}$, a joint reference for the robot lower limbs is computed through an inverse kinematic process within the CartesI/O framework~\cite{cartesio}. The \textit{Robot control} node exploits also XBot2~\cite{XBot2}, in charge of communicating with the robot.

For the experimental setup of this work with the CENTAURO robot, $\boldsymbol{K}$ is set equal to zero. Additionally, the last row of the $\boldsymbol{K}_{cart,cp}$ matrix has been set to zero to limit the mobility ability of the robot along the $z$ axis.

\subsection{Wearable sensorimotor interface}
\begin{figure}
    \centering	\includegraphics[width=0.8\linewidth]{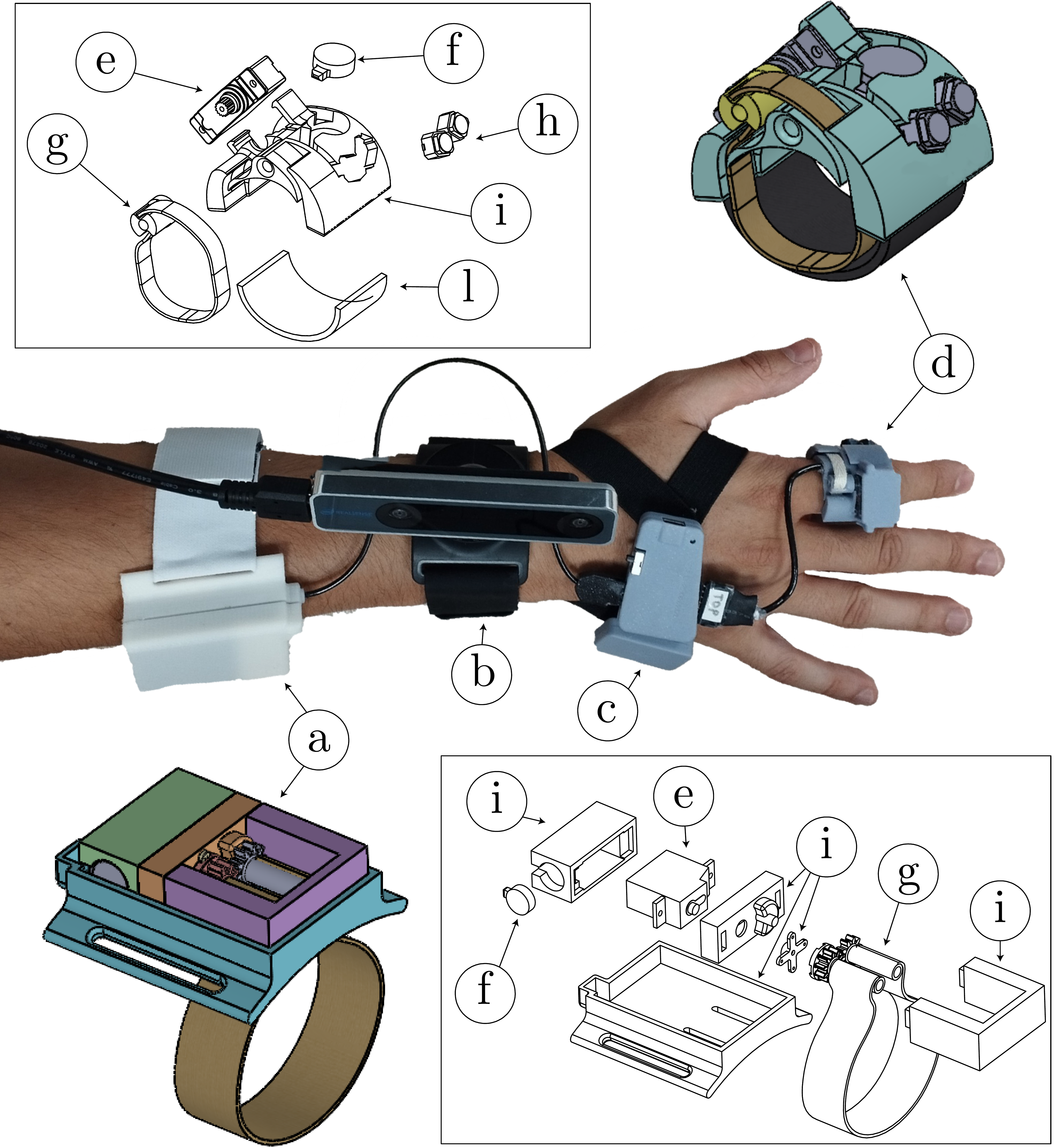}
	\caption{Sensorimotor interface: (a) forearm haptic interface, (b) tracking camera, (c) PCB and battery, (d) ring haptic interface, (e) servomotor, (f) vibromotor, (g) folding mechanism, (h) buttons, (i) structural plastic parts, (l) elastic band. \label{fig:hapticinterface}
     \vspace{-20px}}
\end{figure}

The adopted wearable sensorimotor interface sends inputs to the TPO system through tracking cameras and buttons and gives haptic feedback to the user in the form of vibratory and skin indentation stimuli. The user wears a tracking camera on each wrist, an armband on each forearm, and a ring on each index finger (\fref{fig:hapticinterface}). Cameras (Intel\textsuperscript{\textregistered} RealSense T265) are used to track the operator's arms movement as explained in \sref{sec:TPO}, while all other components are a novelty of this work. Both the armbands and the ring can deliver skin indentation stimuli through a fabric belt that is folded and unfolded by a servomotor (both Hitech servomotors, models HS-5035HD and HS-53), and vibrations produced by coin vibration motors (Precision Microdrives Model No. 310-103.004, 10mm Vibration Motor). The rings are also endowed with two push buttons each (Alpsalpine SPEH110100). 
The control board is a custom-designed board based on the Espressif ESP. We built it to ensure a stable wireless connection and to physically connect all the components needed to program the microcontroller, drive the motors, read the buttons, and recharge the battery in a small amount of space. The result is a lightweight interface: \SI{114}{\gram} for the camera and \SI{104}{\gram} for the rest.

The working principle of the skin indentation devices is based on~\cite{pacchio-hring}, with some improvements. In particular, as shown in \fref{fig:hapticinterface}, both the forearm and the ring haptic devices use only one motor to drive the fabric belt: in the forearm this has been implemented with opposing gears, while in the ring with a winch mechanism. 
The structure has been designed to be compact and adaptable to the physical characteristics of different users. The forearm haptic interface is adjustable with a Velcro fastener, and the ring has an interchangeable elastic band.

The mapping between haptic stimuli and delivered information is summarized in \tref{tab:map_feedback}. When commanding the robot with a particular arm, the user experiences a squeezing (indentation) sensation on the forearm generated by the worn armband proportionally to the magnitude of the applied virtual force. 
In the previous works on TPO, the user could not feel the resistance caused by the application of a virtual force on the robot, which constituted a difference with respect to the application of an actual force during a physical human-robot interaction. With this work, instead, the human-robot connection is physically perceived by the user through the haptic feedback, while still keeping the human in a remote position.
Skin indentation is also implemented in the rings to deliver information on the robot interaction with its surroundings. To account for both, prehensile and non-prehensile manipulation actions, we envisaged a setup in which the right arm of the robot was endowed with a gripper (the DAGANA, a 1-DOF beak-like gripper) and the left one with a passive ball-shaped end-effector. %
The user's right finger was squeezed based on the grasping force applied by the gripper on a grasped object, while the left finger was squeezed proportionally to the contact force between the robot left end-effector and the environment.

\begin{table}
	\centering
	\caption{Mapping of the haptic feedback}
	\label{tab:map_feedback}
	\begin{tabularx}{\linewidth}{l c c}
		
		\toprule
		 & Squeeze Feedback & Vibration Feedback \\
		\midrule
		
		\textbf{R. Forearm} & R. virtual force magnitude & R. toggle activation ACK \\
		\textbf{R. Finger} & Gripper grasping force & Toggle gripper ACK \\
		\textbf{L. Forearm} & L. virtual force magnitude & L. toggle activation ACK \\
		\textbf{L. Finger} & \scriptsize{L. EE external force magnitude} & Arm/base \textit{cp} change ACK \\
		\bottomrule
	\end{tabularx}
\end{table}

Another improvement introduced in this work is the addition of input buttons embedded in the rings. In~\cite{TPO}, the operator did not have the possibility of independently controlling certain functionalities, like activating and deactivating the teleoperation, changing the control point (e.g., command the robot base instead of the arm), or opening/closing the robot gripper. The solution therein adopted was the inclusion of a second operator who executed these additional commands from a computer according to the first operator's vocal instructions. 
In the new system, thanks to push buttons, the presence of the second operator is not necessary anymore. Furthermore, exploiting the functionalities of the developed haptic devices, two different vibration patterns on the forearm and on the fingers are generated to acknowledge (ACK) the operator about the correct execution of the functionality associated with the pressed button. 
A single short vibration signaled the engagement of the functionality\replaced{;}{,} conversely, a double short vibration signaled the disengagement of the functionality. For example, referring to \tref{tab:map_feedback}, the \virgolette{toggle activation ACK} vibrates once when the user activates the teleoperation, and vibrates twice when the user deactivates the teleoperation. %

It is worth noticing that the mapping between signals and feedback stimuli generation and between additional inputs and buttons has been chosen according to what was deemed more intuitive considering the control paradigm and the employment of the interface in a loco-manipulation mission.
For example, since the user commands the robot through his/her arms\added{ with the virtual ropes attached to his/her wrists}, the feedback related to the virtual force is\added{ co-located near the wrist, and} delivered to the forearms\replaced{. Instead,}{, whereas} since humans usually interact with the environment with their hands\added{ and fingers}, the stimuli related to the robot-environment interaction are delivered to the rings.
\replaced{Nevertheless, the software architecture is flexible and different mapping choices can be made, which is an evaluation that we plan for future works.}{Nevertheless, the software architecture is flexible and different choices can be made. In this work, we are not evaluating the numerous amount of possible mappings since it is out of the scope, but we plan to evaluate all these possibilities in future developments.}

%% file: sez04_experiments.tex
\begin{figure}
	\centering
	\includegraphics[width=0.85\linewidth]{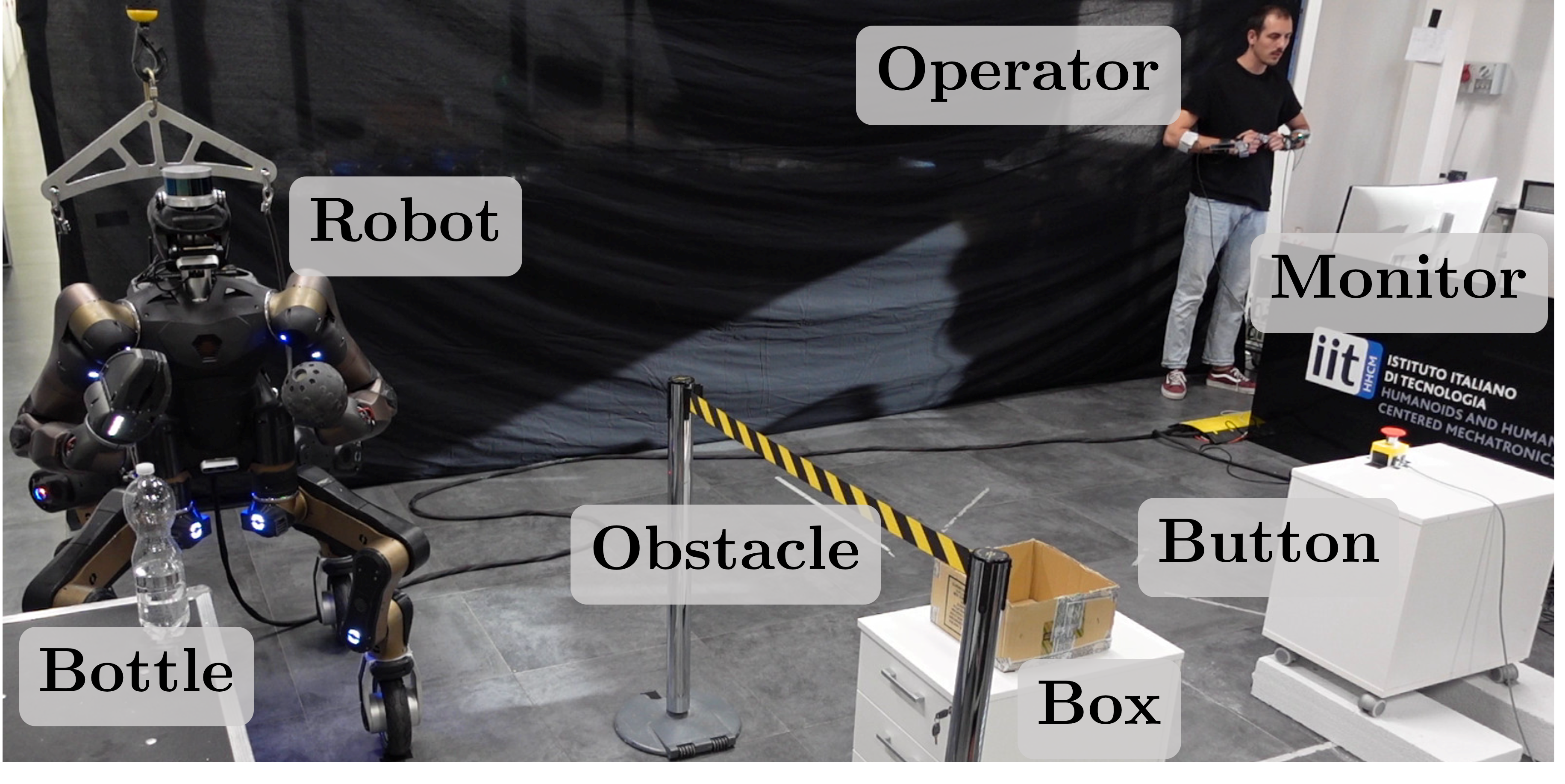}\\
    \includegraphics[width=0.8\linewidth]{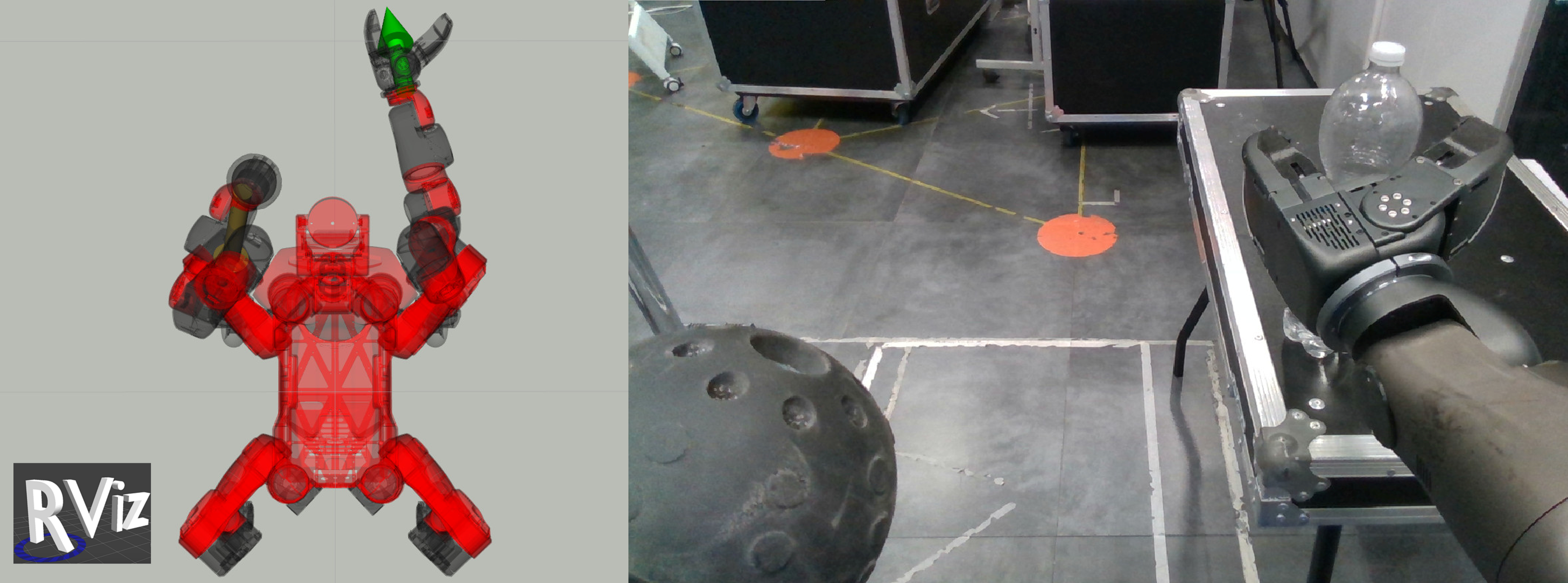}
	\caption{Setup: (top) experimental area, (bottom) operator's monitor with RViz showing virtual forces on the robot model (left) and the robot camera view (right).}
	\label{fig:setup}
\end{figure}

The experimental setup is shown in \fref{fig:setup},
where the operator controls the robot standing in front of a monitor. The monitor, as shown in the bottom part of \fref{fig:setup}, shows the CENTAURO head camera output and an RViz environment with the robot twin and arrows representing the virtual forces applied to the robot parts. %
Users can look at the monitor or at the robot itself at any time, as they prefer.
During the experiments, users control the CENTAURO robot manipulation (left/right arms and the right gripper) and wheeled mobility abilities. 

The mission involves a series of manipulation tasks placed at different locations to be reached sequentially.
In the beginning, the robot is in front of a table where a bottle must be picked up with the right gripper. Then, avoiding an obstacle, the robot must be guided to another location to place the bottle inside a box. Lastly, the robot must be moved close to an emergency button that must be pressed with the left end-effector. We consider a single task failed if the bottle falls off and if the robot hits an object (the obstacle, the table, or one of the drawers supporting the box and the button), but we let the rest of the mission continue (if the bottle drops, we manually place it in the robot gripper to let continue with the box task). The time of the mission starts as soon the operator activates the teleoperation and ends when the emergency button at the last location is pressed.

The user controls the robot generating virtual forces as explained in \sref{sec:TPO}.
We have limited the possible control points to the right/left end-effectors and the robot base. The right arm of the user is always connected to the robot right end-effector, while the left arm of the user can switch between the robot left end-effector and the robot base.
Besides generating virtual forces, the user has $4$ additional input commands to activate/deactivate the control (one for each arm), open/close the gripper, and switch the left arm control point. The way these inputs are commanded depends on the condition (\textit{A}, \textit{B}, \textit{C}) used in the experimental comparison.

\textbf{Condition \textit{A}.} The operator has no haptic feedback, nor push buttons and necessitates a second operator to command the $4$ additional inputs described above. The second operator gives vocal acknowledgments after s/he executes the requested command (\enquote{Ok} word). This condition corresponds to the method presented in our previous work~\cite{TPO}.

\textbf{Condition \textit{B}.} The operator can use the push buttons but does not receive haptic feedback.

\textbf{Condition \textit{C}.} The operator has the full wearable interface available, including buttons and haptic feedback.

\tref{tab:map_feedback} and \tref{tab:map_inputs} summarize the mapping of tactile feedback signals and of the $4$ additional inputs, respectively.

\begin{table}
	\centering
	\caption{Mapping of the inputs for the $3$ modalities}
	\label{tab:map_inputs}
	\setlength{\tabcolsep}{4pt} %
	\begin{tabularx}{\linewidth}{l c c}
		
		\toprule
		& Vocal Input (A) & Button Input (B, C)\\ 
		\midrule
		{\scriptsize \textbf{Right virtual force activation}} & \enquote{Right} & Right button 1    \\
		{\scriptsize \textbf{Left virtual force activation}} & \enquote{Left} & Left button 1   \\
		{\scriptsize \textbf{Open/Close gripper}} & \enquote{Gripper} & Right button 2    \\
		{\scriptsize \textbf{Left EE/Base control point change}} & \enquote{Change} & Left button 2   \\
		\bottomrule
	\end{tabularx}
\end{table}

The experiments involved $12$ participants, $10$ males, $2$ females, with age ranging from $25$ to $39$, 
voluntarily participating after having signed an informed consent. Each experiment lasted about $2$ hours and consisted in a \textit{training phase} followed by a \textit{testing phase}. The training phase lasted around $40$ minutes during which the experimenter taught the participant how to control the robot and let the participant try the mission in the 3 different conditions. Furthermore, the device \replaced{minimum and maximum squeezing force have}{squeezing range has} been set up according to the user sensibility and physical characteristics\added{, such as the forearm and finger circumferences}. 
In the testing phase, each participant executed the whole mission $6$ times, $2$ for each condition. The order of the conditions has been permuted following a certain sequence for each participant (e.g., \textit{ABC}, \textit{ACB}, \textit{BAC}, etc.). Immediately after the two trials for each condition, participants were asked to fill in the NASA-TLX questionnaire referred to the tested condition. At the end of all the $6$ trials, participants assigned the weights to the NASA-TLX factors, and compiled two additional questionnaires. The first questionnaire was about the wearable interface and was composed of $16$ questions formulated as $5$-level Likert items with answers varying
from \virgolette{Strongly disagree} to \virgolette{Strongly agree}, similarly to~\cite{casalino2018operator}. The second questionnaire compared the three conditions one-vs-one (i.e., \textit{A}vs\textit{B}, \textit{A}vs\textit{C}, \textit{B}vs\textit{C}) and was formulated as a 7-point linear scale going from a certain condition to another, similarly to~\cite{maderna2022flexible}. While the statements in the first questionnaire were customized to the specific application presented in this paper, the statements in the linear scale included the whole \replaced{System Usability Scale (SUS)}{SUS}~\cite{brooke1996sus} and two additional questions taken from the \replaced{Usefulness, Satisfaction, and Ease of
use (USE)}{USE} questionnaire~\cite{lund2001measuring}.

Highlights of the experiments are shown in the video attached to this paper\added{, also available at \href{https://youtu.be/QLU2ZrU0HjQ}{https://youtu.be/QLU2ZrU0HjQ}}; we also provide the raw videos of all the trials at\deleted{ the following link:} \href{https://youtu.be/VTiB6I_fIWo}{https://youtu.be/VTiB6I\_fIWo}.

%% file: sez05_discussion.tex
\begin{table}
	\centering
	\caption{Experimental results}
	\label{tab:results}
	\begin{tabularx}{\linewidth}{l c c c c}
		\toprule
            & \multicolumn{2}{c}{Completion Time [s]} & Task & NASA-TLX\\
		  & All & Last Two Trials & Failures & OW\\
		\midrule
		\textbf{\textit{A}} & $168.5\pm28.0$ & $150.1\pm28.1$ & 2 & $42.0\pm17.4$ \\
		  \textbf{\textit{B}} & $178.5\pm2.1$ & $172.5\pm13.4$ & 1 & $47.25\pm19.1$\\
            \textbf{\textit{C}} & $182.6\pm17.2$ & $148.2\pm6.0$ & 1 & $43.02\pm17.6$\\
		\bottomrule
	\end{tabularx}
\end{table}

\begin{figure}
	\centering
	\includegraphics[width=0.9\linewidth]{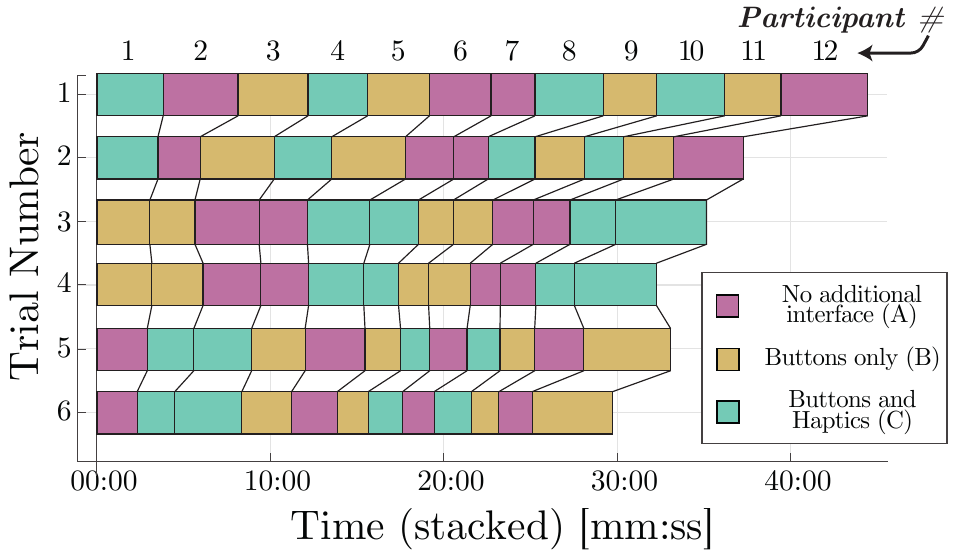}
	\caption{Times of each trial for all the participants, where the conditions have been highlighted in different colors.}
	\label{fig:times}
	\vspace{-5px}
\end{figure}

Obtained results about the completion time averaged among all participants and about the number of total failures are reported in \tref{tab:results}. 
While no statistical significance was found among the completion time values in the different conditions (using a one-way repeated measures ANOVA, $p > .05$), we observed that considering all trials and all users, independently \replaced{of}{from} the condition, a clear trend emerges: the cumulative completion time improved from the first to the last trial, as shown in \fref{fig:times}. This is why in \tref{tab:results} also the completion time employed in the last two trials is reported. Here we can see that the condition \textit{C} performs slightly better than the other two, but still without statistical significance. %
The occurred failures are only a few and happened while grasping the bottle or in cases where the robot collided with other objects in the scene. %

\begin{figure}
	\centering
	\includegraphics[width=0.9\linewidth]{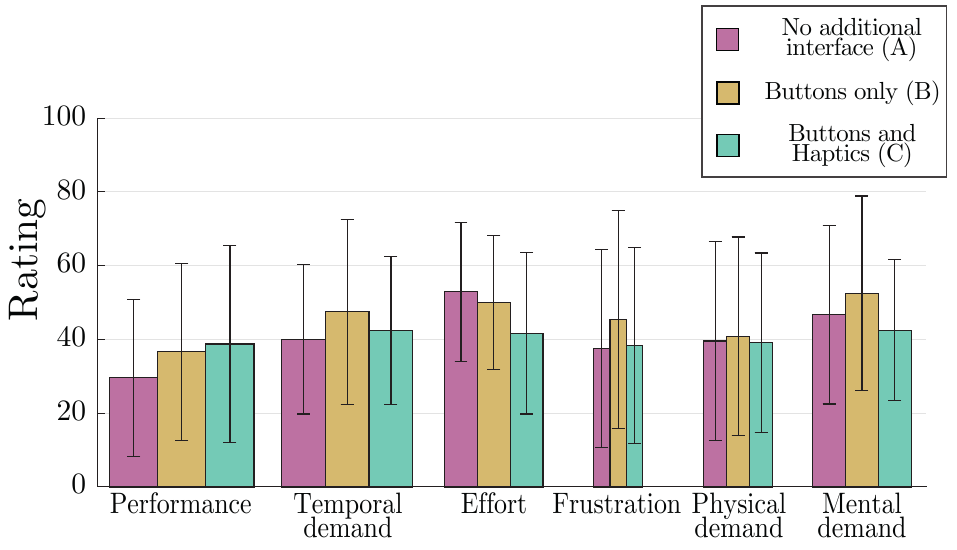}
	\caption{NASA-TLX Ratings.\deleted{We can put the legend on a row if we need more space}}
	\label{fig:nasatlx}
	\vspace{-10px}
\end{figure}

The results for the single weighted categories of NASA-TLX are plotted in \fref{fig:nasatlx} according to the guidelines in~\cite{hart1988development}, while the overall workload (OW) is shown in the last column of \tref{tab:results}. 
There was no statistically significant difference between the means of the ratings provided for the different interfaces (one-way repeated measures ANOVA, $p > .05$). The fact that all conditions show comparable results in terms of workload indicates that adding an extra feedback channel with respect to the visual one does not generate an additional burden on the users. %

\begin{figure}
	\centering
	\includegraphics[width=1\linewidth]{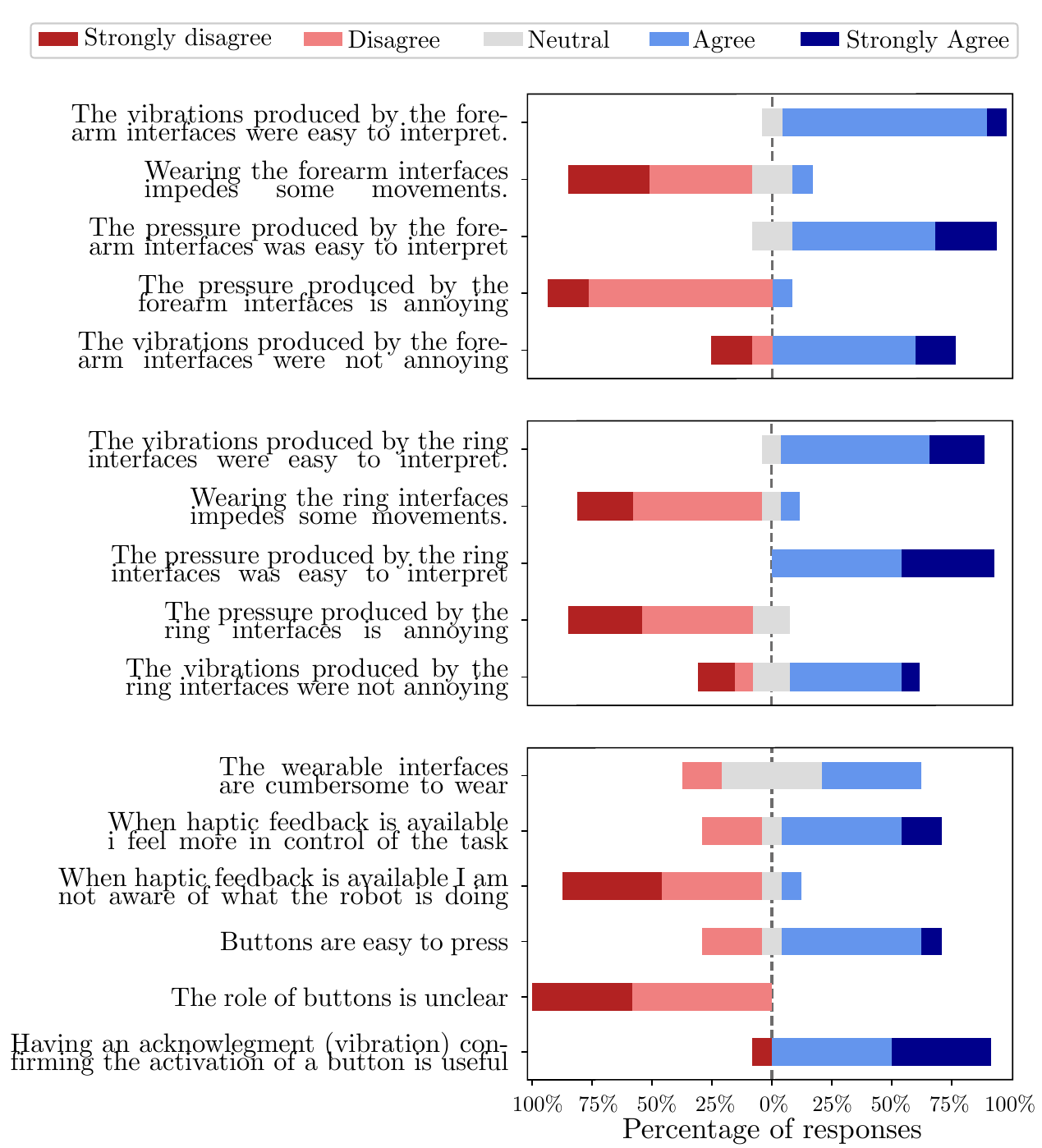}	
    \caption{Likert questionnaire \replaced{about}{on} the haptic devices.}
	\label{fig:likert}
\end{figure}

\begin{figure}
	\centering
	\includegraphics[width=1\linewidth]{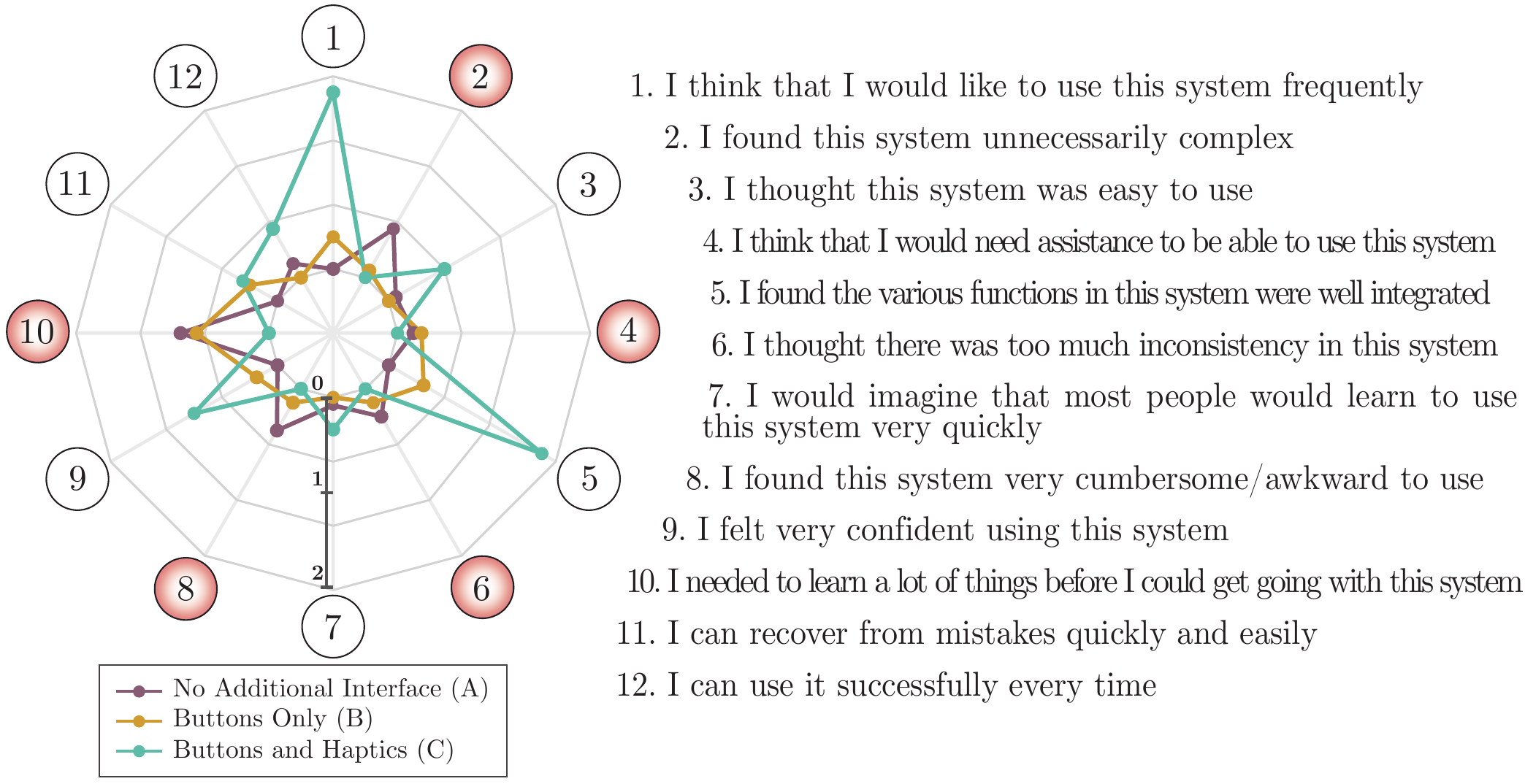}	
    \caption{Scores gathered from the second questionnaire. The outer ring represents a score of $2$ for each question. Questions in red have been reversed to consider always a high score as a positive rating for a condition.}
	\label{fig:avsbvsc_spider}
\end{figure}

The results of the two additional questionnaires are shown in \fref{fig:likert} and \fref{fig:avsbvsc_spider}, respectively.
In the first one, users assessed the features of the single devices, forearm and ring, and of the whole haptic interface. The results reflect positive assessments of the haptic interface, with users commending its effectiveness and the benefits of incorporating such haptic stimuli in the interface.
From the results of the second questionnaire, a score has been extracted, resulting in the plot of \fref{fig:avsbvsc_spider}. 
Considering the 7-point linear scale going from a certain condition to the other for each question of the three one-vs-one comparisons, no score has been assigned if the participant selected the middle point. For the positive questions ($1$, $3$, $5$, $7$, $9$, $11$, $12$) a score of $+1$, $+2$ or $+3$ is given to one of the conditions selected depending on the point chosen. On the contrary, for the negative questions ($2$, $4$, $6$, $8$, $10$), the score of $+1$, $+2$ or $+3$ has been given to the condition opposed to the selected point, to always consider a spike in the spider plot as a positive rating. For each question, the resulting score shown in the plot is the average among all participants.

In general, we can notice the preferences for the condition \textit{C} against the two others, (questions $1$, $9$, $11$, $12$), considering it only slightly more complicated (questions $2$, $3$, $7$, $8$), hence feeling the necessity to \replaced{spend}{have} more time to learn it (questions $4$, $10$), but judging the haptic interface well integrated (questions $5$, $6$). Interestingly, the condition with only the buttons available (\textit{B}) is not always preferred over the condition without the additional interface (\textit{A}). This is also visible in the NASA-TLX overall workload of \tref{tab:results}. Also according to users' free answers, some of them preferred the vocal commands of condition \textit{A} since they did not have to use their hands, nor to learn the buttons mapping. It is important to notice that the vocal commands were directed to a human operator who, in all cases, perfectly understood the commands and rapidly reacted to them. In the same situation, a speech recognition software might have been less responsive and accurate. Moreover, the second operator always gave an auditory feedback after the execution of a command, which instead is not present in the condition \textit{B}. A vibration feedback after the pressing of a button is instead present in the condition \textit{C}, which indeed has been considered very important to have by the participants. Anyway, the conditions \textit{B} and \textit{C} have both the evident advantage of not \replaced{requiring}{necessitating} the second operator to execute the additional inputs.

%% file: sez06_conclusion.tex
In our previous work, we introduced the concept of TelePhysicalOperation, a kind of teleoperation interface, that combines the intuitiveness of a physical human-robot interaction with the safety of controlling a robot remotely, through virtual ropes attached to selected robot body parts, like in a \enquote{Marionette} kind of interface, while receiving only visual feedback from the robot. This work presented an evolution of the TPO framework, which integrates additional sensing and feedback functionalities by means of a wearable haptic interface, composed of forearm and ring devices to be worn on each arm of the user.
We have developed haptic devices to generate different kinds of feedback according to the user's inputs to the robot.
By mapping the forearm squeezing feedback on the virtual forces applied with the TPO interface, the users have a direct sensation about the virtual forces impressed on the robot, consisting of the \enquote{Marionette} metaphor. Squeezing forces on the index fingers, instead, alert the user about external forces experienced by the robot end-effectors. Additional input buttons have been added to the ring device to expand the user's possible inputs to be given to the robot, avoiding the previously necessary vocal interaction with a second operator.

Twelve participants have been involved in validating the TPO way of control both with and without the proposed haptic interface. The mission consisted in a series of tasks where the manipulation and locomotion ability of the CENTAURO robotic platform had to be exploited. 
Results show a positive outcome for the devices integrated in the interface. Furthermore, comparing \replaced{the conditions without haptic feedback versus the haptics-enabled TPO}{the haptics-enabled TPO with the conditions without haptic feedback}, results show a general subjective preference for the \replaced{latter}{first}, considering only a slight increase in complexity to wear the enhanced interface. 
In the future, we plan to conduct a more in-depth study of the single feedback and its effect on such kinds of teleoperation interfaces, as well as to validate and compare other potential mapping configurations between the sensorimotor input/feedback functionalities and the robot actions and interactions.

%% file: main.bbl
\begin{thebibliography}{10}
\providecommand{\url}[1]{#1}
\csname url@rmstyle\endcsname
\providecommand{\newblock}{\relax}
\providecommand{\bibinfo}[2]{#2}
\providecommand\BIBentrySTDinterwordspacing{\spaceskip=0pt\relax}
\providecommand\BIBentryALTinterwordstretchfactor{4}
\providecommand\BIBentryALTinterwordspacing{\spaceskip=\fontdimen2\font plus
\BIBentryALTinterwordstretchfactor\fontdimen3\font minus
  \fontdimen4\font\relax}
\providecommand\BIBforeignlanguage[2]{{%
\expandafter\ifx\csname l@#1\endcsname\relax
\typeout{** WARNING: IEEEtran.bst: No hyphenation pattern has been}%
\typeout{** loaded for the language `#1'. Using the pattern for}%
\typeout{** the default language instead.}%
\else
\language=\csname l@#1\endcsname
\fi
#2}}

\bibitem{Selvaggio2021}
M.~Selvaggio, M.~Cognetti, S.~Nikolaidis, S.~Ivaldi, and B.~Siciliano,
  ``Autonomy in physical human-robot interaction: A brief survey,''
  \emph{{IEEE} Robot. Autom. Lett.}, vol.~6, no.~4, pp. 7989--7996, 2021.

\bibitem{TPO2}
D.~Torielli, L.~Muratore, and N.~Tsagarakis, ``Manipulability-aware shared
  locomanipulation motion generation for teleoperation of mobile
  manipulators,'' in \emph{{IEEE/RSJ} Int. Conf. Intell. Robots Syst.}, 2022,
  pp. 6205--6212.

\bibitem{Casadio2012}
\BIBentryALTinterwordspacing
M.~Casadio, R.~Ranganathan, and F.~A. Mussa-Ivaldi, ``The body-machine
  interface: A new perspective on an old theme,'' \emph{J. Motor Behavior},
  vol.~44, no.~6, pp. 419--433, 2012. [Online]. Available:
  \url{https://doi.org/10.1080/00222895.2012.700968}
\BIBentrySTDinterwordspacing

\bibitem{Gasper2021}
\BIBentryALTinterwordspacing
G.~{\v{S}}kulj, R.~Vrabi{\v{c}}, and P.~Podr{\v{z}}aj, ``A wearable {IMU}
  system for flexible teleoperation of a collaborative industrial robot,''
  \emph{Sensors}, vol.~21, no.~17, 2021. [Online]. Available:
  \url{https://doi.org/10.3390/s21175871}
\BIBentrySTDinterwordspacing

\bibitem{TPO}
\BIBentryALTinterwordspacing
D.~Torielli, L.~Muratore, A.~Laurenzi, and N.~Tsagarakis,
  ``Telephysicaloperation: Remote robot control based on a virtual
  “marionette” type interaction interface,'' \emph{{IEEE} Robot. Autom.
  Lett.}, vol.~7, no.~2, pp. 2479--2486, 2022. [Online]. Available:
  \url{https://doi.org/10.1109/LRA.2022.3144792}
\BIBentrySTDinterwordspacing

\bibitem{pacchierotti2015cutaneous}
C.~Pacchierotti, D.~Prattichizzo, and K.~J. Kuchenbecker, ``Cutaneous feedback
  of fingertip deformation and vibration for palpation in robotic surgery,''
  \emph{IEEE Trans. Biomed. Eng.}, vol.~63, no.~2, pp. 278--287, 2015.

\bibitem{dafarra2022}
\BIBentryALTinterwordspacing
S.~Dafarra, K.~Darvish, R.~Grieco, G.~Milani, U.~Pattacini, L.~Rapetti,
  G.~Romualdi, M.~Salvi, A.~Scalzo, I.~Sorrentino, D.~Tomè, S.~Traversaro,
  E.~Valli, P.~M. Viceconte, G.~Metta, M.~Maggiali, and D.~Pucci, ``{iCub3}
  avatar system,'' 2022. [Online]. Available:
  \url{https://doi.org/10.48550/arXiv.2203.06972}
\BIBentrySTDinterwordspacing

\bibitem{prattichizzo2021human}
D.~Prattichizzo, M.~Pozzi, T.~L. Baldi, M.~Malvezzi, I.~Hussain, S.~Rossi, and
  G.~Salvietti, ``Human augmentation by wearable supernumerary robotic limbs:
  review and perspectives,'' \emph{Prog. Biomed. Eng.}, vol.~3, no.~4, p.
  042005, 2021.

\bibitem{Franco2019}
L.~Franco, G.~Salvietti, and D.~Prattichizzo, ``Command acknowledge through
  tactile feedback improves the usability of an emg-based interface for the
  frontalis muscle,'' in \emph{IEEE World Haptics Conf.}, 2019, pp. 574--579.

\bibitem{Centauro2}
N.~Kashiri \emph{et~al.}, ``{CENTAURO}: A hybrid locomotion and high power
  resilient manipulation platform,'' \emph{{IEEE} Robot. Autom. Lett.}, vol.~4,
  no.~2, pp. 1595--1602, 2019.

\bibitem{Gaofeng2023}
\BIBentryALTinterwordspacing
G.~Li, Q.~Li, C.~Yang, Y.~Su, Z.~Yuan, and X.~Wu, ``The classification and new
  trends of shared control strategies in telerobotic systems: A survey,''
  \emph{IEEE Trans. Haptics}, vol.~16, no.~2, pp. 118--133, 2023. [Online].
  Available: \url{https://doi.org/10.1109/TOH.2023.3253856}
\BIBentrySTDinterwordspacing

\bibitem{VILLANI2018}
\BIBentryALTinterwordspacing
V.~Villani, F.~Pini, F.~Leali, and C.~Secchi, ``Survey on human–robot
  collaboration in industrial settings: Safety, intuitive interfaces and
  applications,'' \emph{Mechatronics}, vol.~55, pp. 248--266, 2018. [Online].
  Available: \url{https://doi.org/10.1016/j.mechatronics.2018.02.009}
\BIBentrySTDinterwordspacing

\bibitem{nunez2018}
\BIBentryALTinterwordspacing
{M. N{\'{u}}{\~{n}}ez L.}, D.~Dajles, and F.~Siles, ``Teleoperation of a
  humanoid robot using an optical motion capture system,'' in \emph{IEEE Int.
  Work Conf. Bioinspired Intell.}, 2018, pp. 1--8. [Online]. Available:
  \url{https://doi.org/10.1109/IWOBI.2018.8464136}
\BIBentrySTDinterwordspacing

\bibitem{Vogel2011}
J.~Vogel, C.~Castellini, and P.~van~der Smagt, ``Emg-based teleoperation and
  manipulation with the dlr lwr-iii,'' in \emph{{IEEE/RSJ} Int. Conf. Intell.
  Robots Syst.}, 2011, pp. 672--678.

\bibitem{Darvish23}
\BIBentryALTinterwordspacing
K.~Darvish, L.~Penco, J.~Ramos, R.~Cisneros, J.~Pratt, E.~Yoshida, S.~Ivaldi,
  and D.~Pucci, ``Teleoperation of humanoid robots: A survey,'' \emph{IEEE
  Trans. Robot.}, vol.~39, no.~3, pp. 1706--1727, 2023. [Online]. Available:
  \url{https://doi.org/10.1109/TRO.2023.3236952}
\BIBentrySTDinterwordspacing

\bibitem{MONIRUZZAMAN2022}
\BIBentryALTinterwordspacing
M.~Moniruzzaman, A.~Rassau, D.~Chai, and S.~M.~S. Islam, ``Teleoperation
  methods and enhancement techniques for mobile robots: A comprehensive
  survey,'' \emph{Robot. Auton. Syst.}, vol. 150, p. 103973, 2022. [Online].
  Available: \url{https://doi.org/10.1016/j.robot.2021.103973}
\BIBentrySTDinterwordspacing

\bibitem{Dargahi2004}
\BIBentryALTinterwordspacing
J.~Dargahi and S.~Najarian, ``Human tactile perception as a standard for
  artificial tactile sensing—a review,'' \emph{Int. J. Med. Robot. Comput.
  Assisted Surgery}, vol.~1, no.~1, pp. 23--35, 2004. [Online]. Available:
  \url{https://doi.org/10.1002/rcs.3}
\BIBentrySTDinterwordspacing

\bibitem{Pacchierotti2015}
\BIBentryALTinterwordspacing
C.~Pacchierotti, L.~Meli, F.~Chinello, M.~Malvezzi, and D.~Prattichizzo,
  ``Cutaneous haptic feedback to ensure the stability of robotic teleoperation
  systems,'' \emph{Int. J. Robot. Res.}, vol.~34, no.~14, pp. 1773--1787, 2015.
  [Online]. Available: \url{https://doi.org/10.1177/0278364915603135}
\BIBentrySTDinterwordspacing

\bibitem{Zhao2023}
\BIBentryALTinterwordspacing
L.~Zhao, M.~Nybacka, and M.~Rothh{\"{a}}mel, ``A survey of teleoperation:
  Driving feedback,'' in \emph{IEEE Intell. Vehicles Symp.}, 2023, pp. 1--8.
  [Online]. Available: \url{https://doi.org/10.1109/IV55152.2023.10186553}
\BIBentrySTDinterwordspacing

\bibitem{XIA2023}
\BIBentryALTinterwordspacing
P.~Xia, F.~Xu, Z.~Song, S.~Li, and J.~Du, ``Sensory augmentation for subsea
  robot teleoperation,'' \emph{Computers Industry}, vol. 145, p. 103836, 2023.
  [Online]. Available: \url{https://doi.org/10.1016/j.compind.2022.103836}
\BIBentrySTDinterwordspacing

\bibitem{Macchini2020}
\BIBentryALTinterwordspacing
M.~Macchini, T.~Havy, A.~Weber, F.~Schiano, and D.~Floreano, ``Hand-worn haptic
  interface for drone teleoperation,'' in \emph{IEEE Int. Conf. Robot. Autom.},
  2020, pp. 10\,212--10\,218. [Online]. Available:
  \url{https://doi.org/10.1109/ICRA40945.2020.9196664}
\BIBentrySTDinterwordspacing

\bibitem{Schwarz2021}
\BIBentryALTinterwordspacing
M.~Schwarz, C.~Lenz, A.~Rochow, M.~Schreiber, and S.~Behnke, ``Nimbro avatar:
  Interactive immersive telepresence with force-feedback telemanipulation,'' in
  \emph{{IEEE/RSJ} Int. Conf. Intell. Robots Syst.}, 2021, pp. 5312--5319.
  [Online]. Available: \url{https://doi.org/10.1109/IROS51168.2021.9636191}
\BIBentrySTDinterwordspacing

\bibitem{baek2022}
\BIBentryALTinterwordspacing
D.~Baek, Y.~Chen, Chang, and J.~Ramos, ``A study of shared-control with force
  feedback for obstacle avoidance in whole-body telelocomotion of a wheeled
  humanoid,'' 2022. [Online]. Available:
  \url{https://doi.org/10.48550/arXiv.2209.03994}
\BIBentrySTDinterwordspacing

\bibitem{pacchierotti2017wearable}
C.~Pacchierotti, S.~Sinclair, M.~Solazzi, A.~Frisoli, V.~Hayward, and
  D.~Prattichizzo, ``Wearable haptic systems for the fingertip and the hand:
  taxonomy, review, and perspectives,'' \emph{IEEE Trans. Haptics}, vol.~10,
  no.~4, pp. 580--600, 2017.

\bibitem{Brygo2014}
\BIBentryALTinterwordspacing
A.~Brygo, I.~Sarakoglou, N.~Garcia-Hernandez, and N.~Tsagarakis, ``Humanoid
  robot teleoperation with vibrotactile based balancing feedback,'' in
  \emph{Haptics: Neuroscience, Devices, Modeling, and Applications}, 2014, pp.
  266--275. [Online]. Available:
  \url{https://doi.org/10.1007/978-3-662-44196-1_33}
\BIBentrySTDinterwordspacing

\bibitem{Bimbo2017}
\BIBentryALTinterwordspacing
J.~Bimbo, C.~Pacchierotti, M.~Aggravi, N.~Tsagarakis, and D.~Prattichizzo,
  ``Teleoperation in cluttered environments using wearable haptic feedback,''
  in \emph{IEEE Int. Conf. Intell. Robots Syst.}, 2017, pp. 3401--3408.
  [Online]. Available: \url{https://doi.org/10.1109/IROS.2017.8206180}
\BIBentrySTDinterwordspacing

\bibitem{BAI2019}
\BIBentryALTinterwordspacing
D.~Bai, F.~Ju, F.~Qi, Y.~Cao, Y.~Wang, and B.~Chen, ``A wearable vibrotactile
  system for distributed guidance in teleoperation and virtual environments,''
  \emph{Proc. Institution of Mech. Engineers, Part H: J. Eng. Medicine}, vol.
  233, no.~2, pp. 244--253, 2019, pMID: 30595086. [Online]. Available:
  \url{https://doi.org/10.1177/0954411918821387}
\BIBentrySTDinterwordspacing

\bibitem{Correa2022}
\BIBentryALTinterwordspacing
M.~Correa, D.~Cárdenas, D.~Carvajal, and J.~Ruiz-del Solar, ``Haptic
  teleoperation of impact hammers in underground mining,'' \emph{Appl.
  Sciences}, vol.~12, no.~3, 2022. [Online]. Available:
  \url{https://www.mdpi.com/2076-3417/12/3/1428}
\BIBentrySTDinterwordspacing

\bibitem{Cheng2022}
J.~Cheng, F.~Abi-Farraj, F.~Farshidian, and M.~Hutter, ``Haptic teleoperation
  of high-dimensional robotic systems using a feedback mpc framework,'' in
  \emph{{IEEE/RSJ} Int. Conf. Intell. Robots Syst.}, 2022, pp. 6197--6204.

\bibitem{cartesio}
A.~Laurenzi, E.~M. Hoffman, L.~Muratore, and N.~Tsagarakis, ``{CartesI/O: A ROS
  Based Real-Time Capable Cartesian Control Framework},'' in \emph{{IEEE} Int.
  Conf. Robot. Autom.}, 2019, pp. 591--596.

\bibitem{XBot2}
A.~Laurenzi, D.~Antonucci, N.~Tsagarakis, and L.~Muratore, ``The xbot2
  real-time middleware for robotics,'' \emph{Robot. Auton. Syst.}, vol. 163, p.
  104379, 2023.

\bibitem{pacchio-hring}
C.~Pacchierotti, G.~Salvietti, I.~Hussain, L.~Meli, and D.~Prattichizzo, ``The
  hring: A wearable haptic device to avoid occlusions in hand tracking,'' in
  \emph{{IEEE} Haptics Symp.}, 2016, pp. 134--139.

\bibitem{casalino2018operator}
A.~Casalino, C.~Messeri, M.~Pozzi, A.~M. Zanchettin, P.~Rocco, and
  D.~Prattichizzo, ``Operator awareness in human-robot collaboration through
  wearable vibrotactile feedback,'' \emph{{IEEE} Robot. Autom. Lett.}, vol.~3,
  no.~4, pp. 4289--4296, 2018.

\bibitem{maderna2022flexible}
R.~Maderna, M.~Pozzi, A.~M. Zanchettin, P.~Rocco, and D.~Prattichizzo,
  ``Flexible scheduling and tactile communication for human-robot
  collaboration,'' \emph{Robot. Computer-Integr. Manuf.}, vol.~73, p. 102233,
  2022.

\bibitem{brooke1996sus}
J.~Brooke, ``{SUS}: a quick and dirty usability scale,'' \emph{Usability Eval.
  Industry}, vol. 189, no.~3, pp. 189--194, 1996.

\bibitem{lund2001measuring}
A.~M. Lund, ``Measuring usability with the {USE} questionnaire,''
  \emph{Usability interface}, vol.~8, no.~2, pp. 3--6, 2001.

\bibitem{hart1988development}
S.~G. Hart and L.~E. Staveland, ``Development of nasa-tlx (task load index):
  Results of empirical and theoretical research,'' in \emph{Advances in
  psychology}.\hskip 1em plus 0.5em minus 0.4em\relax Elsevier, 1988, vol.~52,
  pp. 139--183.

\end{thebibliography}
